\documentclass[letterpaper, 10 pt, conference]{ieeeconf}  

\IEEEoverridecommandlockouts                              

\usepackage{amssymb,amsmath,amsfonts,mathrsfs}
\usepackage{graphicx}
\usepackage[table,xcdraw,dvipsnames]{xcolor}
\usepackage{tikz}
\usepackage{url}

\usepackage[utf8]{inputenc}

\usepackage[labelformat=simple]{subcaption}

\usepackage{colortbl}
\usepackage{booktabs}
\usepackage{array}
\usepackage{tabularx}

\usepackage[flushleft]{threeparttable}
\usepackage{multirow}

\let\labelindent\relax
\usepackage{enumitem}

\newcolumntype{L}[1]{>{\raggedright\let\newline\\\arraybackslash\hspace{0pt}}m{#1}}

\newcommand{\nth}[1]{\ensuremath{#1^\text{th}}}

\makeatletter
\DeclareRobustCommand\onedot{\futurelet\@let@token\@onedot}
\def\@onedot{\ifx\@let@token.\else.\null\fi\xspace}

\usepackage[T1]{fontenc}  
\usepackage[b]{esvect}    

\makeatletter
\newlength\xvec@height%
\newlength\xvec@depth%
\newlength\xvec@width%
\newcommand{\xvec}[2][]{%
  \ifmmode%
    \settoheight{\xvec@height}{$#2$}%
    \settodepth{\xvec@depth}{$#2$}%
    \settowidth{\xvec@width}{$#2$}%
  \else%
    \settoheight{\xvec@height}{#2}%
    \settodepth{\xvec@depth}{#2}%
    \settowidth{\xvec@width}{#2}%
  \fi%
  \def\xvec@arg{#1}%
  \def\xvec@dd{:}%
  \def\xvec@d{.}%
  \raisebox{.2ex}{\raisebox{\xvec@height}{\rlap{%
    \kern.05em
    \begin{tikzpicture}[scale=1]
    \pgfsetroundcap
    \draw (.05em,0)--(\xvec@width-.05em,0);
    \draw (\xvec@width-.05em,0)--(\xvec@width-.15em, .075em);
    \draw (\xvec@width-.05em,0)--(\xvec@width-.15em,-.075em);
    \ifx\xvec@arg\xvec@d%
      \fill(\xvec@width*.45,.5ex) circle (.5pt);%
    \else\ifx\xvec@arg\xvec@dd%
      \fill(\xvec@width*.30,.5ex) circle (.5pt);%
      \fill(\xvec@width*.65,.5ex) circle (.5pt);%
    \fi\fi%
    \end{tikzpicture}%
  }}}%
  #2%
}
\makeatother

\makeatletter
\renewcommand*\env@matrix[1][\arraystretch]{%
  \edef\arraystretch{#1}%
  \hskip -\arraycolsep
  \let\@ifnextchar\new@ifnextchar
  \array{*\c@MaxMatrixCols c}}
\makeatother

\definecolor{commentcolor}{gray}{0.5}
\usepackage{algorithm}
\usepackage{algpseudocode}
\algrenewcommand\algorithmicindent{1.0em}%

\algnewcommand{\LineComment}[1]{\State \textcolor{commentcolor}{\(\triangleright\) #1}}
\algnewcommand{\NewComment}[1]{\textcolor{commentcolor}{\(\triangleright\) #1}}
\algnewcommand{\To}{\textbf{to}}
\algnewcommand{\Break}{\textbf{break}}
\algnewcommand{\Continue}{\textbf{continue}}
\algnewcommand{\IIf}[1]{\State\algorithmicif\ #1\ \algorithmicthen}
\algnewcommand{\EndIIf}{\unskip}
\algnewcommand{\var}[1]{\textit{#1}}
\algnewcommand{\func}[1]{\textsc{#1}}

\newcommand{\figsizetwocol}{0.237}

\title{\LARGE \bf
Efficient Spatial Representation and Routing of Deformable One-Dimensional Objects for Manipulation
}

\def\publishedurl{\centering \textbf{{\color{red}{The published version can be accessed from}} \color{blue}{\url{https://ieeexplore.ieee.org/document/9981939}}}}

\author{Azarakhsh Keipour$^{1}$, 
Maryam Bandari$^{2}$ 
and Stefan Schaal$^{3}$
\thanks{* The publication was written prior to A. Keipour joining Amazon. During the realization of this work, he was affiliated with Carnegie Mellon University and supported by X, The Moonshot Factory residency program.}
\thanks{$^{1}$ Robotics AI, Amazon, Washington, DC
        {\tt\small keipour@gmail.com}}%
\thanks{$^{2,3}$ Intrinsic, Mountain View, CA {\tt\small [maryamb, sschaal] @intrinsic.ai}}%
}

\begin{document}

\maketitle
\thispagestyle{empty}
\pagestyle{empty}

\begin{abstract}

With the field of rigid-body robotics having matured in the last fifty years, routing, planning, and manipulation of deformable objects have recently emerged as a more untouched research area in many fields ranging from surgical robotics to industrial assembly and construction. 
Routing approaches for deformable objects which rely on learned implicit spatial representations (e.g., Learning-from-Demonstration methods) make them vulnerable to changes in the environment and the specific setup. On the other hand, algorithms that entirely separate the spatial representation of the deformable object from the routing and manipulation, often using a representation approach independent of planning, result in slow planning in high dimensional space.

This paper proposes a novel approach to routing deformable one-dimensional objects (e.g., wires, cables, ropes, sutures, threads). This approach utilizes a compact representation for the object, allowing efficient and fast online routing. The spatial representation is based on the geometrical decomposition of the space into convex subspaces, resulting in a discrete coding of the deformable object configuration as a sequence. With such a configuration, the routing problem can be solved using a fast dynamic programming sequence matching method that calculates the next routing move. The proposed method couples the routing and efficient configuration for improved planning time. Our simulation and real experiments show the method correctly computing the next manipulation action in sub-millisecond time and accomplishing various routing and manipulation tasks.

\end{abstract}


\section{Introduction} \label{sec:intro}

\begin{figure}[!t]
\centering
    \begin{subfigure}[b]{\figsizetwocol\textwidth}
        \includegraphics[width=\textwidth, height=4.4cm]{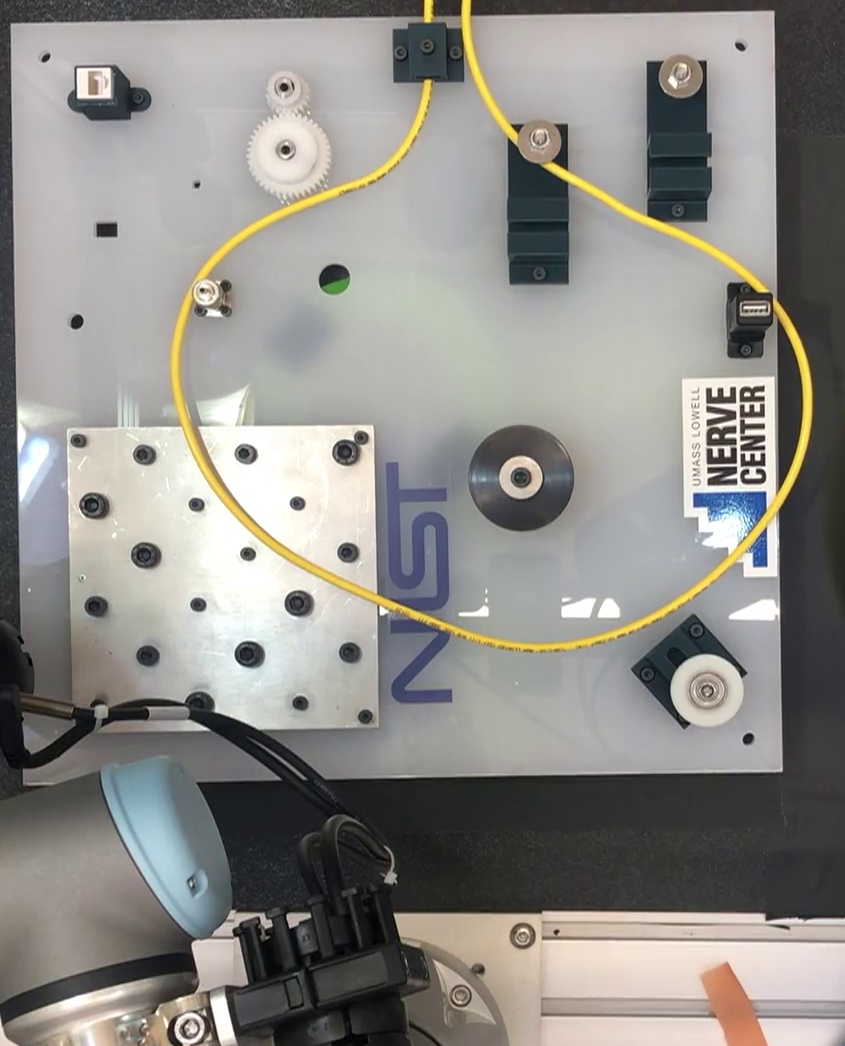}
        \caption{~}
    \end{subfigure}
    \hfill    
    \begin{subfigure}[b]{\figsizetwocol\textwidth}
        \includegraphics[width=\textwidth, height=4.4cm]{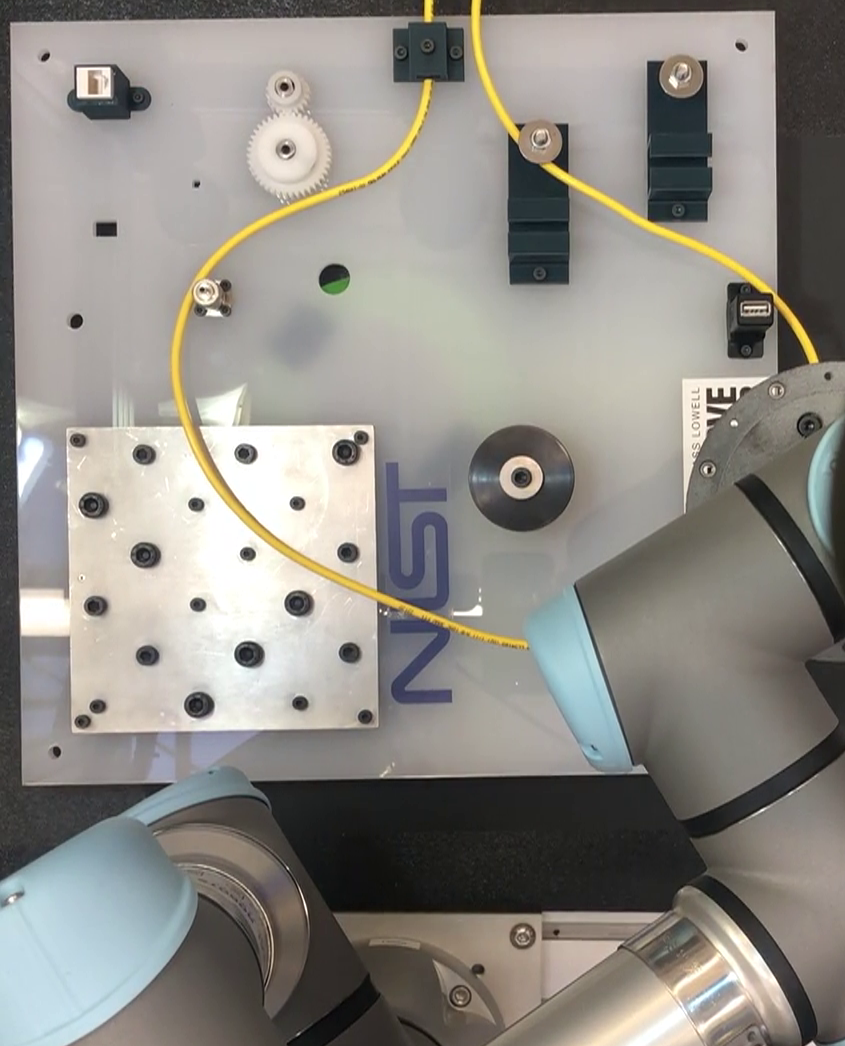}
        \caption{~}
    \end{subfigure}

    \medskip    

    \begin{subfigure}[b]{\figsizetwocol\textwidth}
        \includegraphics[width=\textwidth, height=4.4cm]{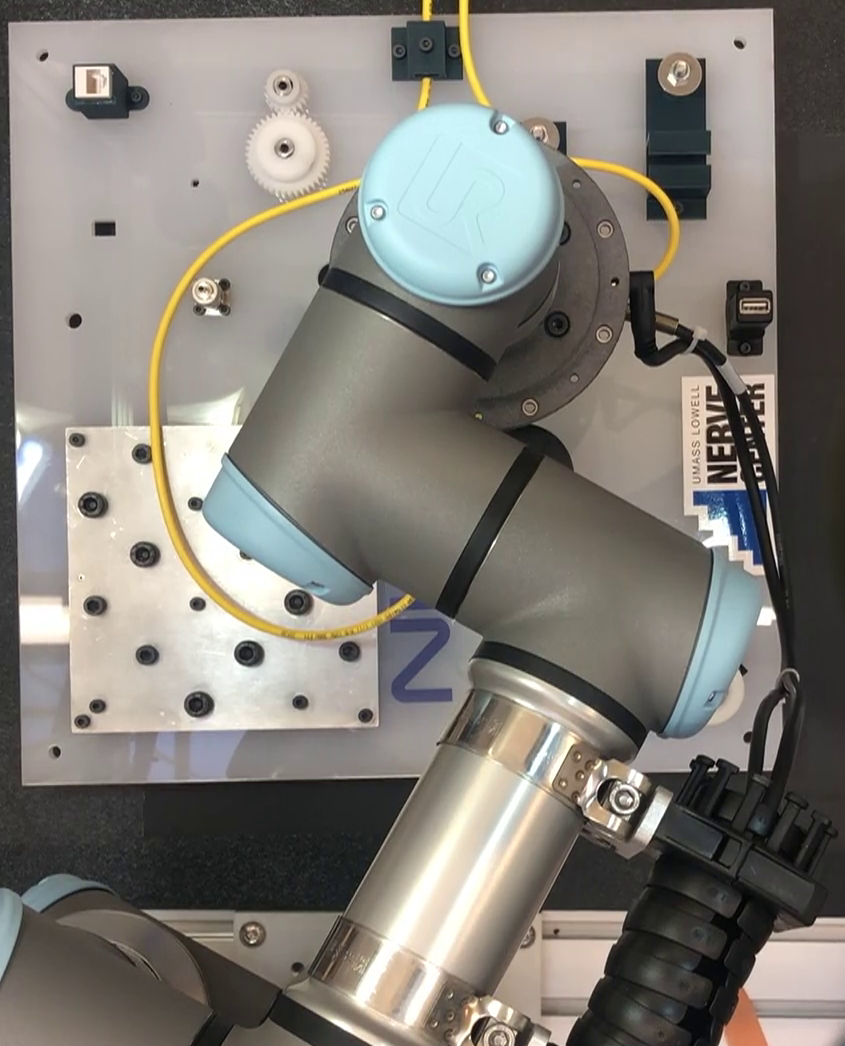}
        \caption{~}
    \end{subfigure}
    \hfill
    \begin{subfigure}[b]{\figsizetwocol\textwidth}
        \includegraphics[width=\textwidth, height=4.4cm]{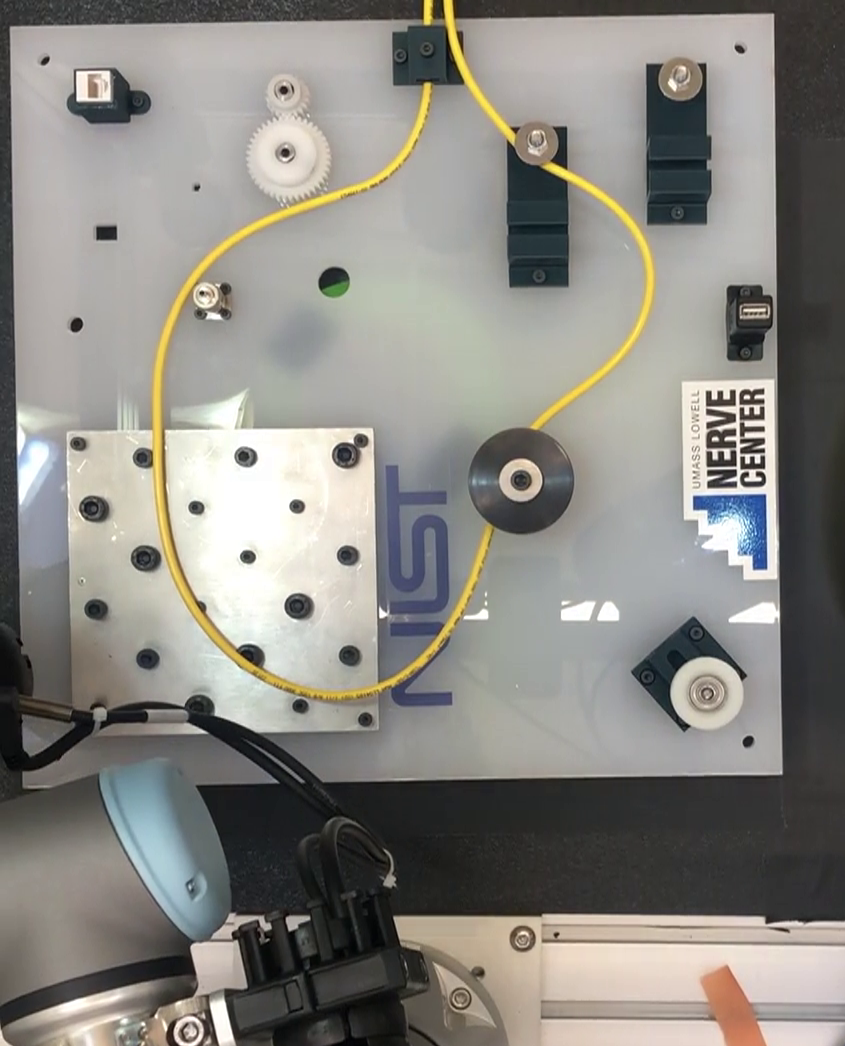}
        \caption{~}
    \end{subfigure}
    
    \medskip
    
    \begin{subfigure}[b]{\figsizetwocol\textwidth}
        \includegraphics[width=\textwidth, height=4.4cm]{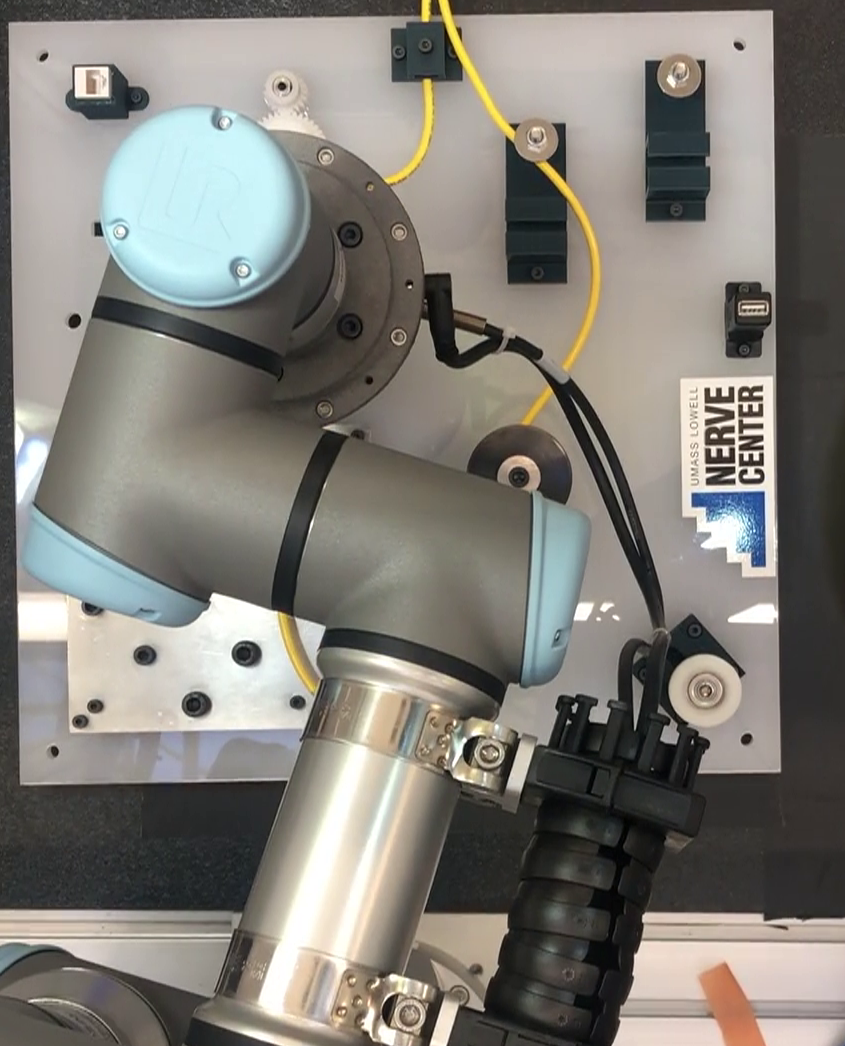}
        \caption{~}
    \end{subfigure}
    \hfill    
    \begin{subfigure}[b]{\figsizetwocol\textwidth}
        \includegraphics[width=\textwidth, height=4.4cm]{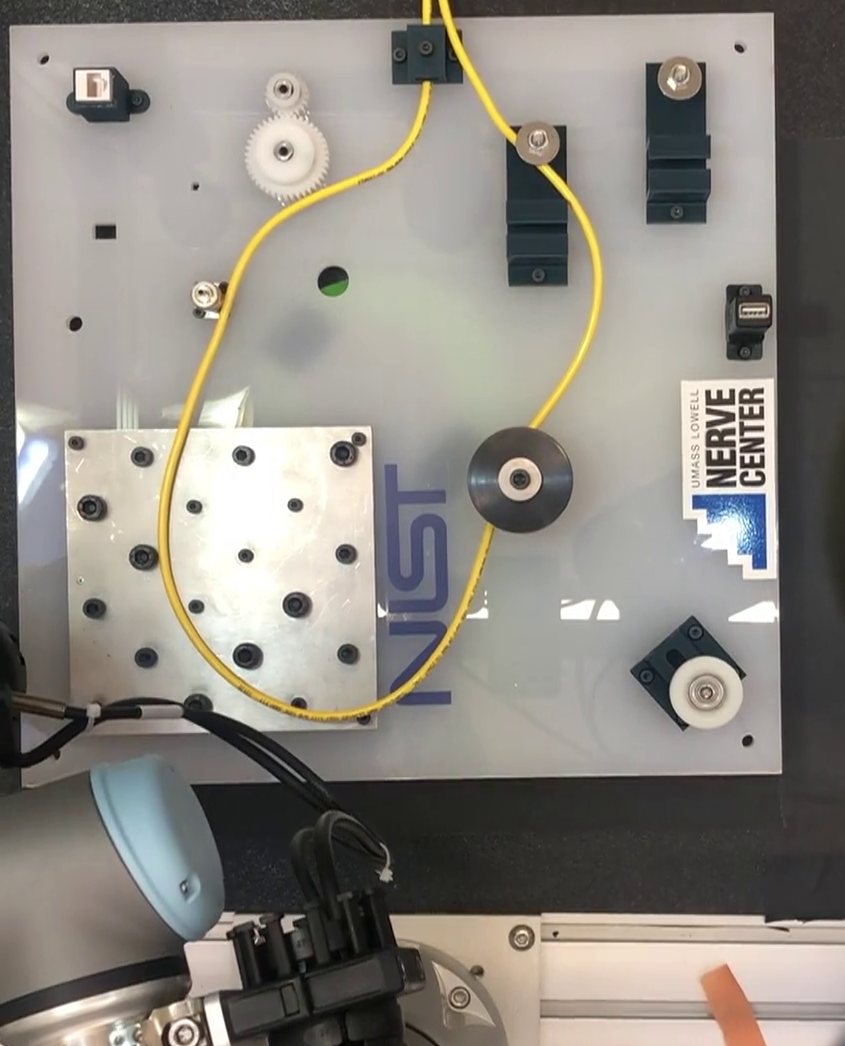}
        \caption{~}
    \end{subfigure}
    \caption{Routing and manipulation of a deformable one-dimensional object from the initial configuration in figure~(a) to the goal configuration in figure~(f). (a) Initial configuration. (b-c) Manipulating the cable based on the first iteration of routing. (d) Resulted configuration after the first round. (e-f) The second iteration of routing and the final result.}
    \label{fig:manipulation}
    \vspace{-4mm}
\end{figure}

Objects such as wires, cables, ropes, threads, and surgical sutures can be found in many industrial, surgical, construction, and everyday settings. In the literature, they are commonly called Deformable One-dimensional Objects (DOOs) or Deformable Linear Objects (DLOs). Automation of tasks involving rigid bodies had been extensively studied in industrial robotics; however, the need for further automation of manual tasks is forcing robotics applications to move towards working with DOOs, raising research interest in various robotics areas~\cite{frobt.2020.00082, 0278364918779698, Keipour:2020:arxiv:integration}. 

A crucial part of many robotics applications involving DOOs is routing~\cite{0278364918779698, GUO2020158, 1236303}. \textit{Routing}, also known as \textit{route planning}, finds a viable \textit{path} to change the initial state of a DOO to a goal state. Path in this context depends on the application and may mean a path in Euclidean space, a trajectory in the configuration space of the DOO, or a series of actions performed on the DOO.

Several representation methods have been used to capture the state of a DOO in route planning and manipulation problems. These representations range from spring-mass models~\cite{LV2017385} to linear combination of curves~\cite{1668249, Hirai2000}, and fixed-length segments~\cite{Keipour:2022:ral:doo, Keipour:2022:thesis}.

Planning methods for routing DOOs for manipulation can be categorized into computational algorithms and approaches based on learning. 

Computational routing methods are generally sampling-based approaches. These methods sample the space and use search methods such as Probabilistic Roadmaps (PRM) and Rapidly-exploring Random Trees (RRTs) for finding the route from the initial DOO state to the final state. Guo et~al.~\cite{GUO2020158} propose the RRT-BwC (Bi-direction with Constrain) planning algorithm to plan for aircraft cables assembly in narrow cabins with obstacles. Their method is based on the geometric formulation of the objects and the bi-directional RRT search to route in the high dimensional planning space. Amato et~al.~\cite{1236303} find an approximate route by pre-computing a global roadmap using a variant of PRM, then refine the route by constrained sampling and applying adaptive forward dynamics. Moll and Kavraki~\cite{1668249} propose DOO planning using minimal-energy curves and a sampling-based planning method such as PRM. Roussel et~al.~\cite{7139627} use quasi-static and dynamic models coupled with sampling-based methods to plan for an elastic rod manipulation. Koo et~al.~\cite{2011jrm} and Ma et~al.~\cite{ma_liu_ding_lv_2020} apply RRT search-based approaches for routing and manipulation of DOOs. For all these methods, the solution can be statistically guaranteed, but they suffer from the curse of dimensionality, and in complex routing cases, the time and space complexity of the algorithms may make the algorithms slow and infeasible for many practical scenarios.

On the other hand, the learning-based methods primarily consist of learning-from-demonstration approaches, potentially working with any DOO type for any task. However, they are not generalizable and quickly fail when the experiment setup conditions even slightly deviate from the learning data~\cite{6386002, 8256194, wang2019learning}.

Our contribution is proposing a novel approach to solving the DOO routing problem that is suitable for both offline and online routing due to its efficiency and speed. This approach relies on the geometrical decomposition of the task space into convex regions. It uses this discretized space to describe the DOO's configuration using a compact sequence, which is simplified from the original 3-D continuous space description. Unlike the existing routing methods that have to find a solution by exploring a high-dimensional space, our new spatial representation allows utilizing a high-speed dynamic programming sequence matching method that reduces the planning delay to near zero, making it suitable for online planning for routing and manipulation tasks.

In the next section, we will describe the problem and our assumptions. Section~\ref{sec:representation} presents our spatial representation method. Section~\ref{sec:routing} proposes our routing approach based on the described spatial representation method. Section~\ref{sec:tests} shows our experiments and presents our results. Finally, Section~\ref{sec:conclusion} discusses the proposed approach and presents the roadmap for further improvements.

\section{Problem Definition} \label{sec:problem}

In the rest of this manuscript, we refer to the area where the task is taking place as \textit{work region}. We assume that the exact positioning of a deformable one-dimensional object is only important inside the work region, and the details of its positioning outside this region are ignored. Additionally, we presume that the work region falls within the workspace of the manipulator robot, and the robot can access all of the work regions. Moreover, we assume that the work region is a "free space" with different \textit{components} occupying some of its space. All the components' positions, shapes, and dimensions are presumed to be known.

The exact positioning of a DOO in the work region is called its \textit{state}, which contains the exact places in space occupied by a DOO. 

Given the initial and desired states of a DOO in a work region, \textit{routing} is finding the sequence of actions required to perform on the DOO to change its state from the initial state to the final state.

The actual DOO state is continuous and needs to be converted to a discrete spatial representation for robotics and computer applications. Common different ways of such representation include sampling equidistant points along the DOO's medial axis, fixed-length B-splines, and fixed-length cylinders connected by spherical joints~\cite{GUO2020158, Keipour:2022:ral:doo, Keipour:2022:icra-workshop:doo}. 

We propose to use a more efficient spatial representation based on the convex decomposition of the work region, combined with a fast sequence matching algorithm to solve the routing problem. Our proposed method is completely independent of the DOO dynamics and tries to embed the dynamics effects in the state representation. We relax the problem assuming that the \textit{slack} of the DOO is not important. In other words, we assume that the application working with the DOO is not affected if the DOO has some extra slack in any area of the work region. For example, suppose a cable is not laying straight in a region and has a rather significant bending in a region. In that case, the extra bending is considered slack, and our algorithm does not consider it. We define slack as extra curves in DOO that do not pass around any components or anchor points; the slack can virtually be eliminated by creating tension in the DOO. 

The following sections describe our proposed spatial representation and the routing method for DOOs.
\section{Spatial Representation} \label{sec:representation}

Let us introduce a graph $G_s$ (called a \textit{spatial representation graph}) to model the spatial representation of a deformable one-dimensional object passing through a work region. 

\begin{figure}[!t]
\centering
    \begin{subfigure}[b]{0.35\textwidth}
        \includegraphics[width=\textwidth]{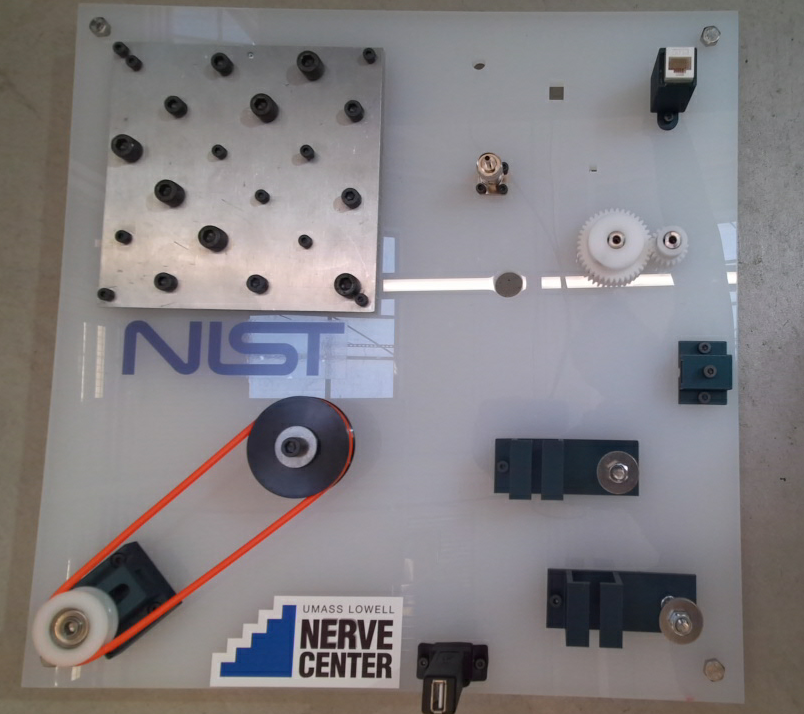}
        \caption{~}
        \label{fig:original-board}
    \end{subfigure}
    ~    
    \begin{subfigure}[b]{0.35\textwidth}
        \includegraphics[width=\textwidth]{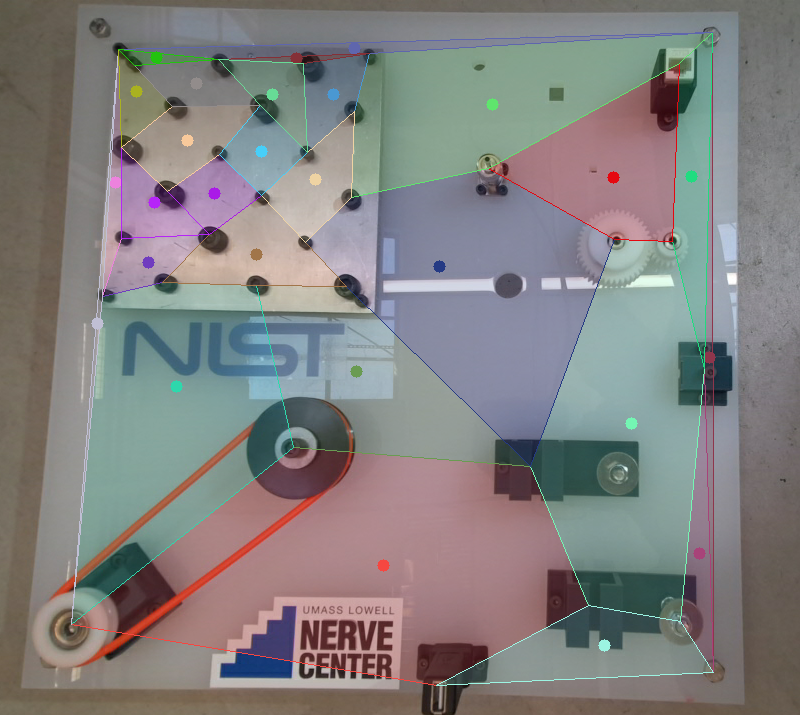}
        \caption{~}
        \label{fig:convex-decomposition}
    \end{subfigure}
    \caption{Convex decomposition of an example circuit board. (a) The original board. (b) The result of convex decomposition. Each component on the board is used for defining the convex region vertices. The centroid of each convex region is marked with a dot.}
    \label{fig:board-regions}
    \vspace{-5mm}
\end{figure}

The work region should be decomposed into convex subspaces to generate the vertices $V_s$ of the spatial representation graph $G_s$. These subspaces are called convex polygons in 2-D and convex polytopes in 3-D spaces. Each of these subspaces is a vertex in the graph $G_s$. If the work region is not enclosed (i.e., if a portion of DOO can lie outside the work region), a new vertex is added to $V_s$ to represent the "outside" region. There are many exact and approximate approaches for convex decomposition, and each can be used for this work~\cite{Deng_2020_CVPR, LIEN2006100, 1236265, 0221025}. Figure~\ref{fig:board-regions} illustrates the convex decomposition on an example circuit board. Each of the components on the board is a node of the convex polygons generated from the board's layout.

\begin{figure}[t]
    \centering
    \includegraphics[width=0.7\linewidth]{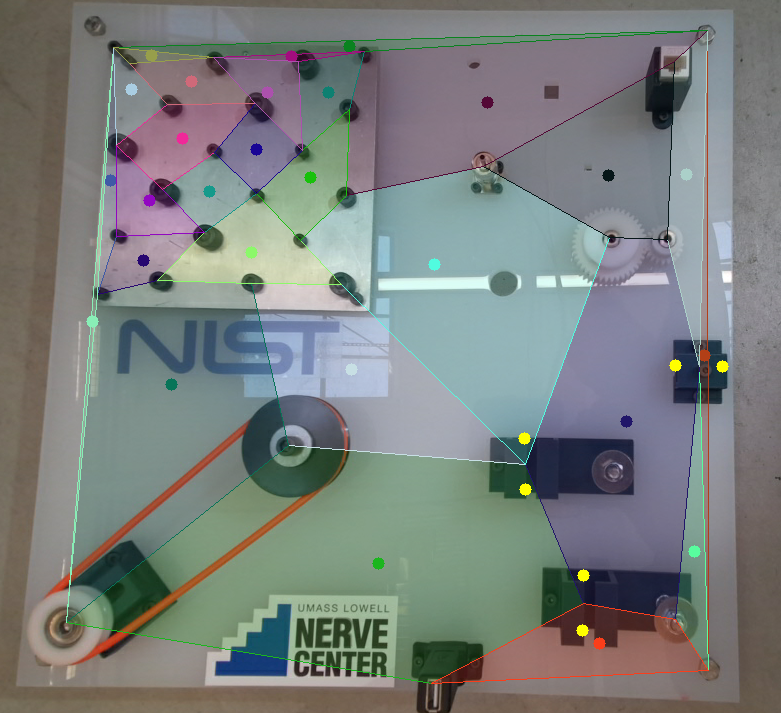}
    \caption{The vertices $V_s$ of the spatial representation graph $G_s$ constructed from the example board in Figure~\ref{fig:original-board}. The tunnel entrance vertices are depicted by yellow dots. Note that the outside region vertex is omitted in the illustration.}
    \label{fig:spatial-graph-vertices}
    \vspace{-3mm}
\end{figure}

The generated convex regions allow efficiently defining subspaces in both 2-D and 3-D. It is desirable to represent the subspace only in 2-D when possible for simplicity. Much of the workspace in the finish line of industrial robotics is on a tabletop which can be approximated as a 2.5-D space. Meaning two dimensions are far more significant than the third dimension. There are specific scenarios where a 3-D work region can be simplified as a 2-D region with additional 3-D "tunnel"-like components such as bridges, passes, and tunnels. To allow the 2-D representation for these work regions, we can add a vertex to $V_s$ for each of the entrances of these components. Figure~\ref{fig:spatial-graph-vertices} shows all the vertices constructed from the example board of Figure~\ref{fig:original-board} with the yellow dots representing the vertices of the entrances of the tunnel components.  

Once all the vertices $V_s$ are defined, the edges $E_s$ for the spatial representation graph $G_s$ can be computed using the following rules:

\begin{itemize}[leftmargin=*]
    \item Vertices from the neighbor convex regions (convex regions sharing a side) are connected with an edge.
    
    \item Convex regions with a side not shared with any other convex region (i.e., convex regions surrounding the work region) are connected to the outside vertex.
    
    \item Vertices for entrances of a tunnel component are connected to each other.
    
    \item Each vertex for the tunnel entrances is connected to the vertex of the convex (or outside) region that it is lying on.
\end{itemize}

\begin{figure}[t]
    \centering
    \includegraphics[width=0.7\linewidth]{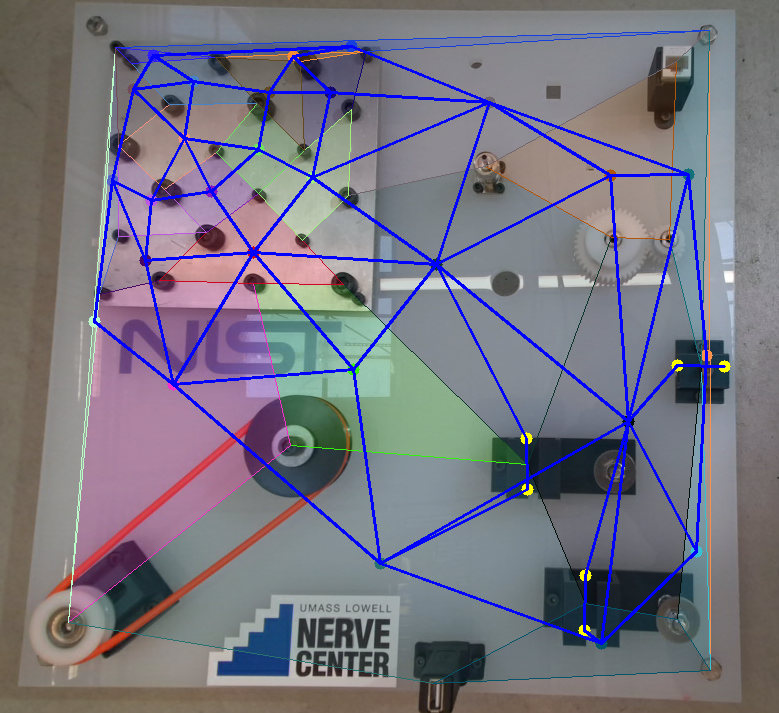}
    \caption{The spatial representation graph $G_s$ computed from the example board of Figure~\ref{fig:original-board}. For simplicity, the edges connected to the outside vertex are not depicted here.}
    \label{fig:spatial-graph}
\end{figure}

Figure~\ref{fig:spatial-graph} shows the spatial representation graph $G_s$ constructed for the circuit board of Figure~\ref{fig:original-board}.

Without the loss of generality, we assume that the size of $V_s$ is $n + 1$, with the vertices numbered from $-1$ to $n - 1$, and $-1$ reserved for the outside vertex. 

\begin{figure}[t]
    \centering
    \includegraphics[width=0.7\linewidth]{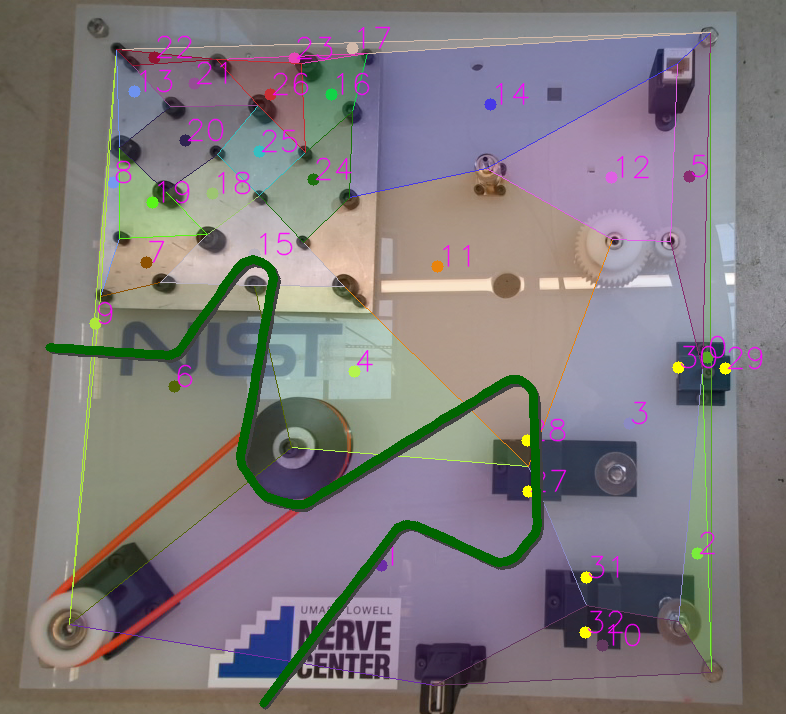}
    \caption{An example DOO passing through the spatial representation graph vertices of Figure~\ref{fig:spatial-graph}.}
    \label{fig:cable-spatial-representation}
    \vspace{-3mm}
\end{figure}

Having computed the graph $G_s$, a DOO lying in the work region or passing through it can be represented by an ordered sequence $C$ of the vertex numbers it is passing through. We call this sequence representing the DOO as \textit{configuration} of DOO. Note that if the DOO is bidirectional (does not have a pre-assigned head and tail), it can have two sequences for the same configuration that are reverse of each other. For example, the configuration of the DOO drawn on the circuit board in Figure~\ref{fig:cable-spatial-representation} is $C = (-1, 1, 27, 28, 11, 4, 1, 6, 4, 15, 6, 9, -1)$ or its reverse. Note that if the graph $G_s$ is computed correctly, every two consecutive vertices in $C$ should have an edge in $E_s$. 

Based on the assumptions of the problem (see Section~\ref{sec:problem}), the extra slack of a DOO in each region is not encoded into its configuration. However, the extra slack is encoded if the DOO goes through some neighboring regions and comes back (e.g., if the DOO "touches" the neighbor convex region while passing through a convex region). Such instances are encoded as palindrome subsequences (i.e., subsequences that are the same if read backward or forward). Removing such subsequences may be desirable depending on the application and simplifies the configuration $C$ of a DOO in the work region.

We should note that the idea of convex decomposition in planning has been explored in other contexts before~\cite{1729881419894787, s19194165, 3022571}. However, it is used differently here for defining the DOO configuration rather than planning itself.

\section{Routing} \label{sec:routing}

Assume that the current configuration $C_0$ of a deformable one-dimensional object in a work region is provided along with the desired goal configuration $C_g$ of the DOO. 

The problem is to route the DOO in the work region from the current configuration $C_0$ to the goal configuration $C_g$. A naive solution to the routing problem is to completely undo $C_0$ into a "free" DOO, then apply $C_g$ configuration by passing through all the vertices in $C_g$. However, this solution is inefficient and requires the maximum number of manipulative actions. A more efficient approach is to keep the matching areas between the current and goal configurations and only manipulate what is necessary to reduce the number of manipulative actions.

We propose utilizing the sequence matching algorithms to minimize the number of actions required to change from $C_0$ into $C_g$. Let us assume that the manipulator supports two motion primitives: 1) pick a DOO at a specific point, and 2) place the picked DOO at a specific point in the work region. Then the following actions on the configuration sequences can be applied:

\begin{itemize}[leftmargin=*]
    \item Replacing the \nth{i} element $s_{i}$ in $C_0$ with the \nth{j} element $g_{j}$ in $C_g$: Pick the DOO where it is passing through vertex $s_{i}$ and place it at vertex $g_{j}$.
    
    \item Removing the \nth{i} element $s_{i}$ in $C_0$ that does not correspond to an element in $C_g$: Remove the DOO from region $s_i$.
    
    \item Inserting the \nth{j} element $g_j$ in $C_g$ that does not correspond to an element in $C_0$: Adding (i.e., stretching) the DOO to region $g_j$.
\end{itemize}

With these three actions, we propose modifying the well-known Levenshtein sequence distance algorithm~\cite{Lev65} to obtain the manipulation actions required for routing. 

The original Levenshtein algorithm computes the minimum required edits (i.e., replacement, deletion, and insertion) to convert the initial sequence to the final sequence. To return this minimum distance, Levenshtein's dynamic programming method computes a matrix that retains the minimum number of edits required for converting the first $i$ elements of the initial sequence to the first $j$ elements of the final sequence. While the algorithm itself only computes the minimum number of edits, the types of edits can be extracted by backtracing this matrix once the algorithm is finished. Note that these edits are not unique, and backtracing will only output only one of the feasible solutions with the minimum number of actions.

To use the Levenshtein algorithm for the routing problem, the following modifications are required:

\begin{enumerate}[leftmargin=*]
    \item When comparing two elements $s_{i}$ and $g_{i}$, they match if they are the same vertex number (i.e., $s_{i}=g_{i}$). However, if they are both $-1$ (the "outside" vertex), they only match if there is a common neighbor for them in the sequence (e.g., if $s_{i-1}=g_{i+1}$). In other words, the two outside regions are considered the same only if they are next to the same vertices. That same vertex may occur before or after $-1$.
    
    \item The cost for each action is set to $1$. However, for a tunnel-like component, the action cost of either of the operations depends on how many vertices come before and after it. In other words, for the \nth{i} element in the sequence of size $n$, the cost will be $2\times\min(i-1, n-i) + 1$. For example, to remove the DOO from a tunnel-like region, it must free either the start of the DOO or the end of the DOO and put everything back again, bypassing the tunnel.
\end{enumerate}



\begin{figure}[!t]
\centering
    \begin{subfigure}[b]{0.36\textwidth}
        \includegraphics[width=\textwidth]{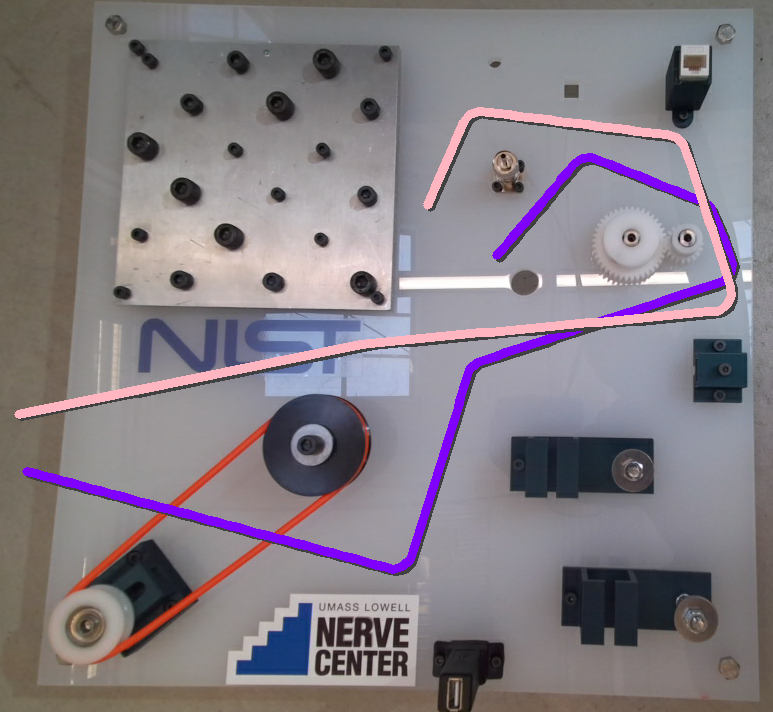}
        \caption{~}
        \label{fig:routing-original}
    \end{subfigure}
    ~    
    \begin{subfigure}[b]{0.36\textwidth}
        \includegraphics[width=\textwidth]{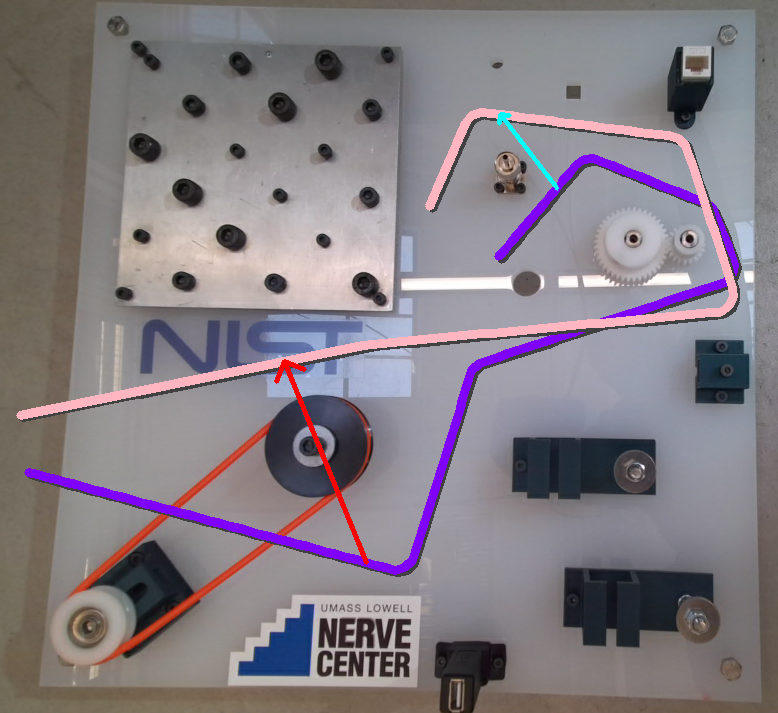}
        \caption{~}
        \label{fig:routing-result}
    \end{subfigure}
    \caption{Routing of a deformable one-dimensional object from the current configuration to a goal configuration. (a) The current (purple) and goal  (pink) deformable objects drawn on the example board of Figure~\ref{fig:original-board}. (b) The result of routing from the current to final configuration. Red and cyan arrows show removal and move (replacement) actions, respectively.}
    \label{fig:routing}
    \vspace{-5mm}
\end{figure}

If the DOO is bi-directional, the algorithm should be repeated with one of the sequences reversed to get the least number of actions. Then, backtracing can give the actions needed to perform on the DOO to change its configuration from $C_0$ to $C_g$. The time and space complexities of the algorithm are $\mathcal{O}(nm)$, where $n$ and $m$ are the lengths of the current configuration ($|C_0|$) and goal ($|C_g|$) configuration sequences. Figure~\ref{fig:routing} shows the routing actions for a DOO to get from its current configuration to the goal configuration.

For realizing the computed actions, a single manipulator can act as below:

\begin{itemize}[leftmargin=*]
    \item For replacement of element $s_{i}$ in $C_0$ with element $g_{j}$ in $C_g$: Pick the DOO where it is passing through vertex $s_{i}$ and place it at vertex $g_{j}$.
    
    \item Removing element $s_{i}$ in $C_0$: Pick the DOO where it is passing through vertex $s_{i}$, and place it at the point where the goal DOO crosses from $g_{j-1}$ to $g_{j}$.
    
    \item Inserting element $g_j$ in $C_g$: Pick the DOO where it is crossing $s_{i-1}$ into $s_i$, and place it at (i.e., stretch it to) vertex $g_{j}$.
\end{itemize}

The details on how these actions are implemented depend on the motion primitives of the robot and the environment. 

We call the point on the DOO where the picking happens as \textit{picking point} and the points where the two sequences match (i.e., no action is required) as \textit{fixed points}. If more than one manipulator is available, the second manipulator can grab the closest fixed point before the picking point, and the third manipulator can grab to the closest fixed point after the picking point to prevent these points from moving. 

The proposed routing algorithm does not incorporate the DOO dynamics and, therefore, cannot understand the result of the actions taken by the manipulator on the whole DOO configuration. To mitigate the lack of dynamics knowledge, routing and manipulation action can be performed iteratively until the current configuration matches the goal configuration. A single routing and then a manipulative action is performed at each iteration to get the DOO closer to the goal configuration. 
\section{Experiments and Results} \label{sec:tests}

To test the proposed method, we implemented it for a routing and manipulation task. The task includes a single-arm manipulating a cable on the circuit board of Figure~\ref{fig:original-board} to change its current configuration to a goal configuration. This board is originally designed for task~\#3 of the Assembly Performance Metrics and Test Methods by the National Institute of Standards and Technology (NIST) to measure the capability of robotics systems for performing advanced manipulation on cables~\cite{nist} and later adopted for the Robotic Grasping and Manipulation Competition in IROS 2020.

We performed many simulation experiments and several experiments with different settings on our robot. Each experiment included several iterations of routing and a manipulative action (i.e., pick and place actions) until the cable configuration had matched the given goal configuration. 

We performed 200 simulation experiments, where we randomly placed a 0.3-0.5~$m$ cable on the 0.38~$m$ NIST board and randomly (in 170 tests) or manually (in 30 tests) placed a cable of the same length on the board as the goal configuration. The average number of actions over all the experiments was 4.34, and the maximum number of actions was 9. The processing time for each routing step was less than 1~$ms$ for all the experiments. Figure~\ref{fig:routing-result} illustrates an iteration of our simulated routing experiments.

For the robot experiments, we manually placed the cables on the NIST board. We considered automated layout detection for this board using real-time rectangle and ellipse detection~\cite{Keipour:2021:ral:ellipse} methods, but used manual specification by the user to save the implementation time. At each step of the planning, we utilized the DOO detection algorithm proposed in~\cite{Keipour:2022:ral:doo} to automatically detect the cable and extract its configuration on the circuit board. Then the goal configuration was manually given to the system. Our experiments showed that the method could also extend to real systems. Figure~\ref{fig:manipulation} shows the routing and manipulation experiments using our Universal Robots UR3 arm robot.

\section{Conclusions and Future Work} \label{sec:conclusion}

We presented a novel method for the spatial representation and routing of a deformable one-dimensional object that is efficient and fast. The proposed method decomposes the work region into convex polygons and polytopes. Then it uses the decomposition to encode the configuration of a DOO in this work region. The resulting configuration is a sequence of the regions the DOO passes from, effectively simplifying the routing algorithm to a modified sequence matching method with a quadratic time and space complexity. The iteration of our routing algorithm with manipulative actions can accomplish the desired routing and manipulation tasks. The low planning time and overhead makes it ideal for offline as well as online planning problems for routing and manipulation. Our experiments showed that the method could correctly plan the manipulation actions and achieve goal configurations of the DOO from various initial configurations. 

The proposed approach is still in its infancy and can be extended further to cover many real-world tasks that are currently being addressed using slower and less efficient methods such as sampling-based planners. In its current iteration, this method can be used in tasks where the environment can be divided to separate convex regions and for a manipulator with two primitive actions: pick a point on the DOO, and place it in a specific point. The algorithm can be easily extended to include more manipulator motion primitives, such as wrapping the DOO around a component or passing through loops to create knots.

Finally, the proposed routing and manipulation algorithm ignores the dynamics of a cable. Although the routing algorithm itself can work well independent from the dynamics, in our real-world tests, we realized that considering the dynamics during manipulative tasks can help the system with performing the pick/place tasks. Additionally, further incorporating simple dynamics into the routing method's cost calculation in the future can reduce the number of actions (i.e., iterations) required for performing a routing/manipulation task by more accurately predicting the result of the manipulation task at each iteration.

\addtolength{\textheight}{0cm}   



\section*{Acknowledgment}

The authors would like to thank Wenzhao Lian for his support and help with robot experiments and implementations.


\bibliographystyle{IEEEtran}
\bibliography{paper-citations.bib}

\end{document}